# Environment-Aware Path Generation for Robotic Additive Manufacturing of Structures

Mahsa Rabiei and Reza Moini*

Civil and Environmental Engineering Department, Princeton University, Princeton, NJ, USA

*Abstract*— **Robotic Additive Manufacturing (AM) has emerged as a scalable and customizable construction method in the last decade. However, current AM design methods rely on pre-conceived (*A priori*) toolpath of the structure, often developed via offline slicing software. Moreover, considering the dynamic construction environments involving obstacles on terrestrial and extraterrestrial environments, there is a need for online path generation methods. Here, an environment-aware path generation framework (PGF) is proposed for the first time in which structures are designed in an online fashion by utilizing four path planning (PP) algorithms (two search-based and two sampling-based). To evaluate the performance of the proposed PGF in different obstacle arrangements (periodic, random) for two types of structures (closed and open), structural (path roughness, turns, offset, Root Mean Square Error (RMSE), deviation) and computational (run time) performance metrics are developed. Most challenging environments (i.e., dense with high number of obstacles) are considered to saturate the feasibility limits of PP algorithms. The capability of each of the four path planners used in the PGF in finding a feasible path is assessed. Finally, the effectiveness of the proposed structural performance metrics is evaluated individually and comparatively, and most essential metrics necessary for evaluation of toolpath of the resulting structures are prescribed. Consequently, the most promising path planners in challenging environments are identified for robotic additive manufacturing applications.**

## I. Introduction

Over the last decade, several robotic additive manufacturing (AM) (also known as 3D printing) approaches have been developed across scales [1] and structures of various complexity [2], [3], [4], [5], [6] using various types of robotic systems such as fixed manipulators [7] and mobile robots [8], [9], [10]. AM approaches to construction, especially on-site construction, are alternatives to conventional techniques which require labor-intensive human involvement and lack customization and ability to handle complexity, especially in unknown environments or outdoor terrains [11]. On-site robotic AM is rapidly growing in construction sectors, demonstrating a pathway for shorter construction time, less taxing labor, and higher customization, such as US Army's on-site Automated Construction of Expeditionary Structures (ACES) sector [12] and NASA's 3D printed extraterrestrial habitats in simulated outdoor environments [13]. Nevertheless, the current AM workflow includes designing the structure in CAD software, converting the design to a 3D surface geometry representation (e.g., .stl format), and slicing the structure into discrete points, using commercial software to generate the toolpath [14]. The output of the workflow is commonly known as G-code [15] in 3D-printers or robotic code [16] (e.g., Rapid code, in case of ABB robotic manipulators).

Although robotic AM is a growing common method for rapid construction of (expeditionary structures on the earth with potential for deployment beyond earth, the current AM workflow cannot handle the unknown nature of the environment, for instance in presence of various obstacles in the workspace or defects in the design. In other words, the path in the current robotic AM frameworks is generated in an offline manner without knowledge of the actual workspace. Therefore, there is a need for an environment-aware (tool) path generation framework that allows for handling the course of the structure with higher degree of complexity and with accounting for obstacles in real-world environments.

Previous research has focused on overcoming the dynamic nature of materials [17], [18], [19], [20], and uncertainties involved in material processing robotic AM, there remains a lack of a framework that addresses the unknowns in the environment in which the structure is constructed. Previous research in path generation has merely generated toolpath using only the conventional path generation workflow, optimized further with regard to variables such as time, length, and quality of the extrusion [21], [22], [23]. In another work, a continuous path generation has been developed for partial and fully-solid infill designs [24]. Similarly, to allow for more complex designs, a path planner for a non-planar structure [25] has been developed to create paths with curvature out of the 2D plane. Additionally, a framework has been proposed that generates the path based on Bezier curves in cases where constant speed in extrusion is required [26]. All current approaches require the A priori design of the structure and do not consider awareness of the environment such as the presence and location of obstacles in the workspace.

In more recent developments, researchers have taken steps towards more intelligent tools for toolpath generation in 3D-printing. Previously established generative design technique that is used in outputting various structural geometries, is proposed by using commercial software considering design parameters such as cost and structural requirements augmented by optimization and machine learning techniques [27]. In other works, a framework, also based on generative design technique, is proposed to generate toolpaths onto arbitrary-shaped 3D substrates based on the depth of the surface [28]. However, the design is prescribed beforehand and also, the obstacles in the environment are not considered. Lastly, a real-time framework augmented by using deep reinforcement learning [29] has recently been proposed in which the robotic AM system scans the exiting structure as it is being built using a camera and completes the remaining structure autonomously. Although this framework is real-time and online, it does not consider the environment or the obstacles within the workspace, i.e., although it is structure-aware, it is not environment-aware.

*Corresponding author: mmoini@princeton.edu)

In this paper, a path generation framework (PGF) is proposed to allow generation of the toolpath for two designs of structure: open (wall with two fixed vertices) and closed (hexagon with six fixed vertices) as shown in Fig. 1 for applications in additive robotic fabrication. The proposed PGF makes use of PP algorithms to generate toolpaths in the presence of obstacles in the workspace (i.e., environment-aware and online).

For applications to robotic AM and to assess the effectiveness of the framework, structural performance metrics are developed and the most promising path planners in challenging environments are identified, accordingly. Insight from this research highlights (i) key performance metric that can sufficiently describe and assess the quality of any path planner for structural and geometric considerations, and (ii) the performance and merit of each path planner of the PGF for use in robotic AM.

Fig. 1. An overview of the environment-aware PGF in (a) open and (b) closed structures

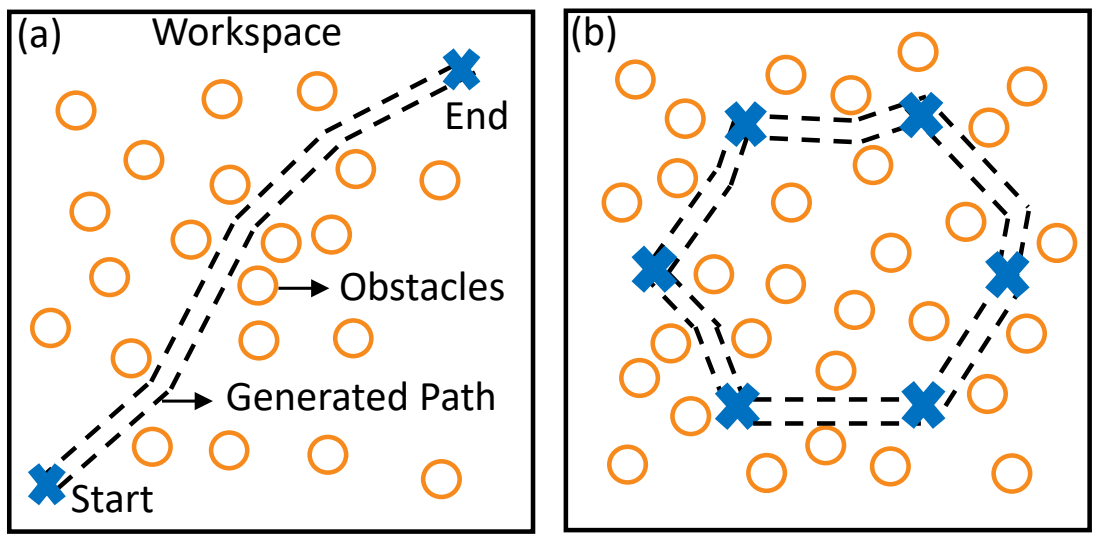

## II. SENSITIVITY ANALYSIS OF KEY PARAMETERS IN PATH PLANNERS

Two main categories for path planners [30], explicit and implicit methods, are explored. Explicit methods include analytical expression for the robot's geometry and dynamics, while implicit methods are data-driven. Although implicit methods are able to handle complex situations, explicit methods are still proven to be the most reliable [30]. Here, we have selected search-based (Dijkstra [31] and A* [32]) and sampling-based (Rapidly-exploring Random Tree (RRT) [33]) and Probabilistic roadmap (PRM) [34]) PP algorithms of explicit methods to employ in the proposed PGF. Each path planner has key parameters that play a vital role in the success rate and run time of finding a feasible path in challenging environments.

To appropriately select the values of the key parameters of the investigated path planners, a series of sensitivity analyses are conducted. As demonstrated in Fig. 2, number of obstacles is the key parameter that affects the success rate of all four path planners. Grid size is a key parameter in search-based path planners such as Dijkstra and A*. Key parameters of PRM include the maximum edge length in which the neighbor nodes can be visited and the number of neighbor nodes to visit to find a feasible path. Number of sample points in the environment is another important parameter in PRM, although it has not been analyzed here as a proper number of sample points along with other parameters is sufficient to find a feasible path. Expansion length of random branches is a key parameter in RRT as it expands randomly in the environment towards the end point. As a note, path resolution is another essential parameter for all path planners. It is the length of path segments to be checked for obstacle collision. Here, this parameter is set equal to the width of the 3D printed filaments.

Fig. 2. Key parameters of (a) Dijkstra and A*, (b) PRM, and (c) RRT

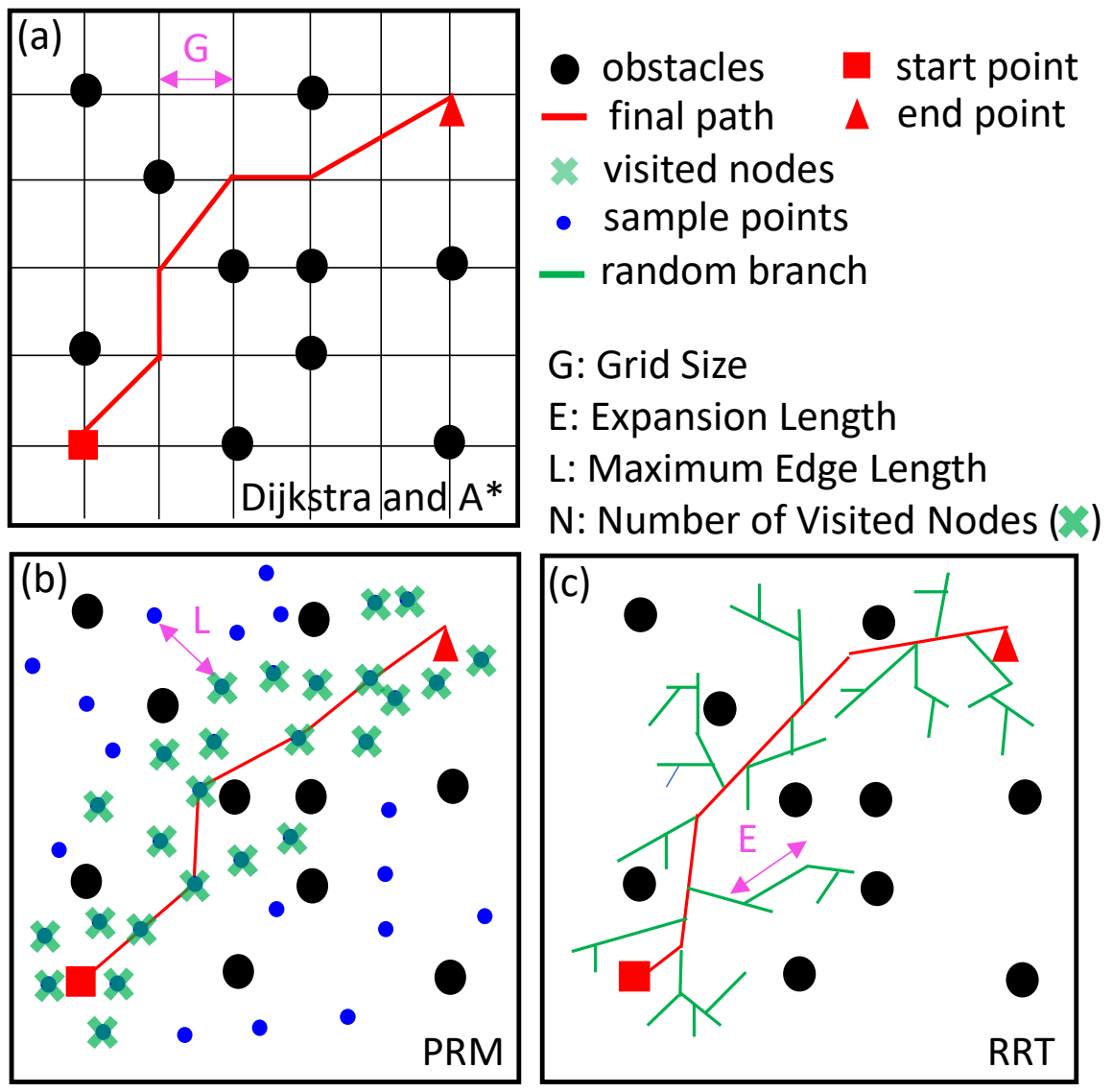

### A. Search-based Path Planning (PP) Algorithms

To determine the number of obstacles, Dijkstra (Fig. 3 (a)) and A* (Fig. 3 (c)) are evaluated in a broad range of obstacles (2 to 1024), assuming the grid size as 10 mm. Generally, the run time rises as the number of obstacles increases. The number of obstacles with 50% success rate is consequently utilized in grid size sensitivity analysis. 128 and 256 obstacles are determined for sensitivity analysis of Dijkstra and A*, respectively. Considering the least number of obstacles between these two values, 128 number of obstacles is selected for further sensitivity analysis of sampling-based path planners.

In addition, a sensitivity analysis is conducted for grid size. As grid size decreases, run time increases. Therefore, greater value of grid size with highest success rate is desired. By evaluating Dijkstra (Fig. 3 (b)) and A* (Fig. 3(d)) in a broad range of grid size (2 to 20mm), 12mm is determined as the maximum grid size with highest success rate and is selected to be utilized in the PGF.

### B. Sampling-based Path Planning (PP) Algorithms

A sensitivity analysis on the expansion length of random branches of RRT is conducted (Fig. 4(a)) in a range of 5 to 400mm, and 20mm is determined as the highest success rate. Unlike grid size in search-based path planners, run time increases as the expansion length in RRT grows.

For PRM, a sensitivity analysis for the number of visited nodes is conducted with a range of 10 to 100 nodes (Fig. 4(b)). As a result, number of visited neighbor nodes is determined as 10 with the highest success rate and shortest run time. Similarly, a sensitivity analysis for maximum edge length (Fig. 4(c)) is conducted with a range of 50 to 800mm. Accordingly, maximum edge length is determined as 410mm.

Fig. 3. Results of sensitivity analysis studies for Dijkstra with respect to (a) number of obstacles and (b) grid size, and for A* with respect to (c) number of obstacles and (d) grid size

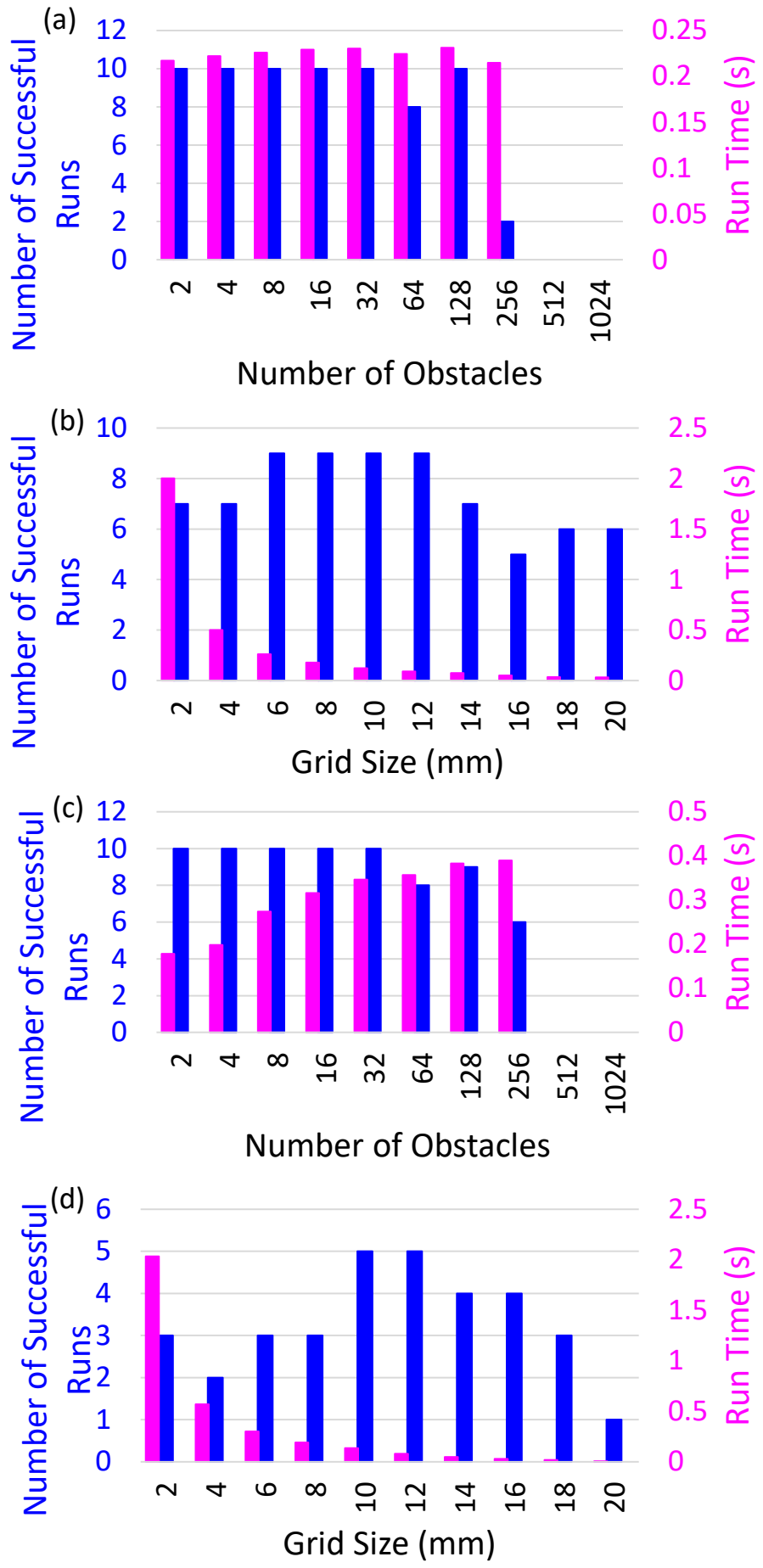

TABLE 1. Results of sensitivity analysis for various key parameters of the four path planners

| | |
|---|---|
| Path Resolution (mm) | 20 |
| Grid Size (Dijkstra and A*) (mm) | 12 |
| Expansion Length (RRT) (mm) | 20 |
| Number of Sample Points (PRM) | 500 |
| Number of Visited Neighbors (PRM) | 10 |
| Maximum Edge Length (PRM) (mm) | 410 |

The final identified values for the key parameters are demonstrated in Table 1.

Fig. 4. Results of sensitivity analysis for (a) expansion length of random branches in RRT, as well as (b) number of visited neighbor nodes in PRM and (c) maximum edge length in PRM

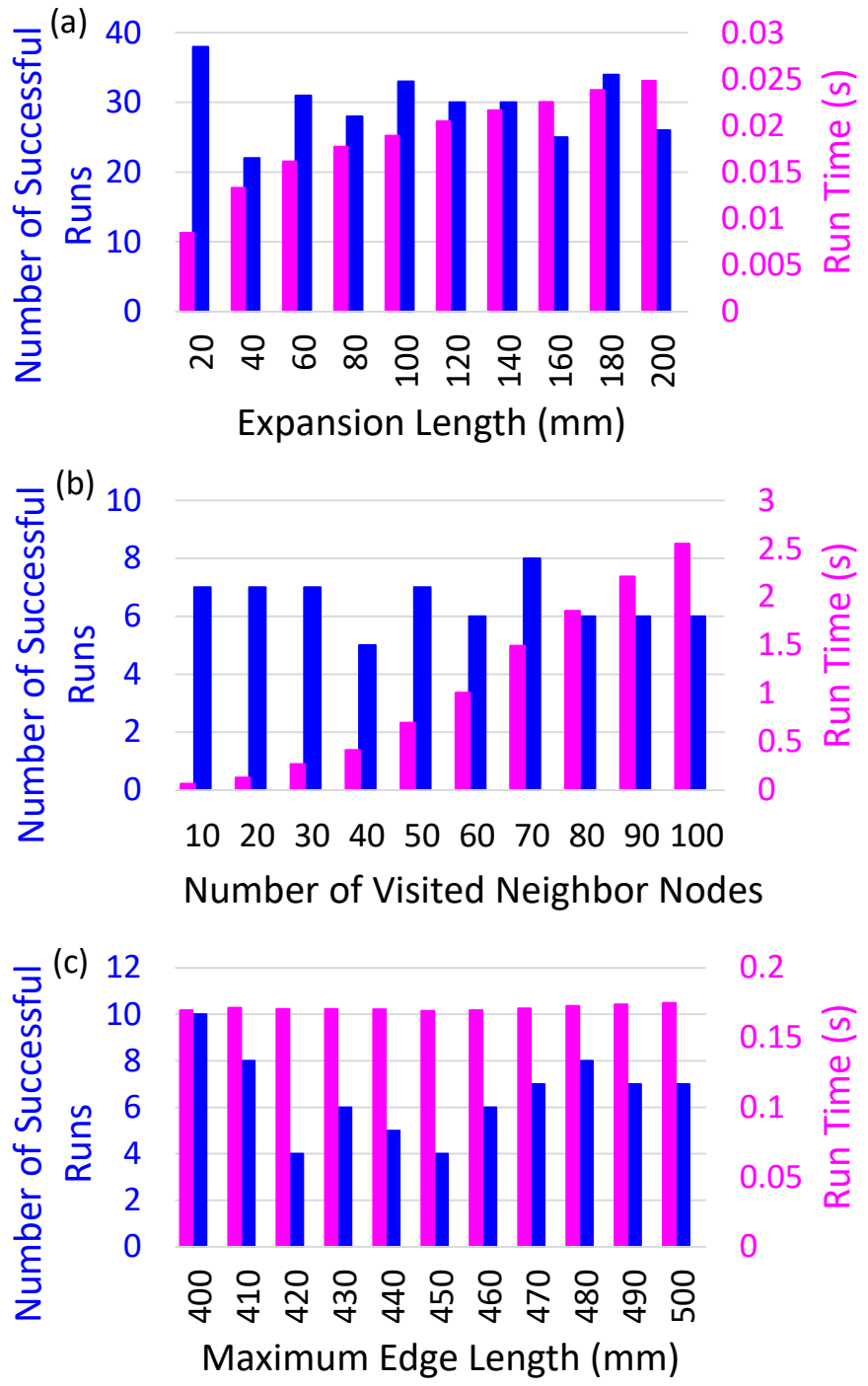

## III. PROPOSED PATH GENERATION FRAMEWORK (PGF)

In the proposed PGF, the fixed vertices are provided by the user. In case of an open structure such as a wall, two vertices including start and end of the wall are considered. In case of a closed structure such as a hexagon, six vertices including the main vertices are assigned. Then, path planners are employed by Python Robotics library [35], [36] in Python to find a feasible path between each pair of vertices. Six performance metrics are considered in the performance assessment metrics (Table 2) to evaluate the generated path in computational (run time) and structural (roughness, number of turns, offset, RMSE, and path deviation) considerations. The structural metrics ensure that the path is smooth with fewer turns, and therefore, it is suitable for robotic AM processes.

Here, the PGF is developed in Python and PP algorithms are employed utilizing the Python Robotics library [35], [36]. In the performance studies presented in the next section, the workspace dimensions of 800 by 800 mm are considered. To generate an open structure, provided start and end vertices as inputs are (100, 100) and (700, 700). For a closed structure, input vertices are (700, 480), (480, 700), (180, 619), (100, 319), (319, 100), and (619, 180). PP algorithms find a feasible path between each pair of vertices, sequentially.

## IV. PERFORMANCE ASSESSMENT STUDIES OF PATH PLANNERS IN THE PROPOSED FRAMEWORK

Although path planners themselves have been generally studied [37], their performance and utility in robotic AM is novel. Here, the performance analysis of path planners for robotic AM of structures is conducted in the most challenging environments to saturate the capacity of path planners and assess their ultimate performance limits.

TABLE 2. Performance assessment metrics (structural/computational)

| | |
|---|---|
| 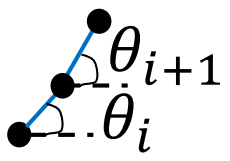 | Roughness (°): average of $\lvert\theta_{i+1}\rvert - \lvert\theta_i\rvert$ for all path segments between each pair of vertices |
| 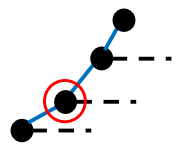 | Number of Turns (-): counts as a turn if $\lvert\theta_{i+1}\rvert - \lvert\theta_i\rvert > 0$ between each consecutive path segments |
| 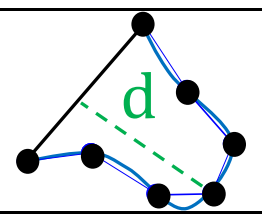  | Offset, d (mm): maximum deviation of path segments from the straight path between each pair of vertices |
| 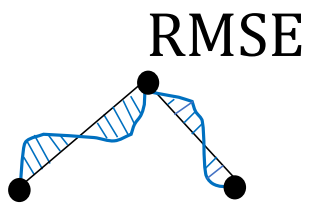  | RMSE (mm): RMSE of deviation of path segments from the straight path between each pair of vertices<br>$\text{RMSE} = \sqrt{\frac{1}{n}\sum_{i=1}^{n}(y_i - \hat{y}_i)^2}$ $y_i$: measured distance; $\hat{y}_i$: ideal distance; $n$: number of segments |
| 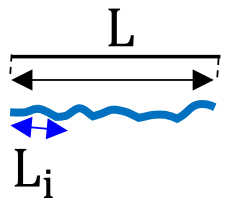  | Path Deviation (-): summation of the length of path segments over the length of the straight path between each pair of vertices<br>Path deviation $= \frac{\sum L_i}{L}$ |
| 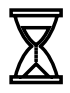 | Run Time (s): required time to find a feasible path between each pair of vertices |

Six performance metrics are measured for each path planner in each environment. The lower the value of the metrics, the higher the corresponding performance.

First, the case of obstacles arranged randomly with different densities are presented. For an open structure with random obstacles, 256 is the highest number of obstacles with which at least one path planner finds a feasible path. In this case, RRT fails, and other path planners determine a path successfully (Fig. 5(a)). RRT is able to find a feasible path in lower obstacle densities. Dijkstra and A* perform similarly, however, they outperform PRM, with ~0.4s shorter run time, ~1mm lower RMSE, ~1.25mm lower offset, and ~22.62° less roughness.

For a closed structure with random obstacles, 64 is the highest number obstacles with which at least one path planner finds a feasible path. In this case, RRT fails, and other three path planners find a path successfully (Fig. 5(b)). A* has the shortest run time (0.75s), lowest RMSE (3.04mm) and offset (4.87mm), although Dijkstra has fewer turns (20) and the least roughness (6.66°). Based on resulting generated paths, roughness and number of turns seem to represent the smoothness more accurately. Therefore, Dijkstra is the smoothest hexagonal pattern. However, if in robotic AM avoiding sharp corners are necessary, PRM performs better, although it has a higher offset and RMSE.

Next, the case of obstacles arranged periodically with different obstacle spacing are presented. It is found that a periodic obstacle arrangement creates a more challenging environment for path planners in terms of finding a feasible path as the internal spacing reduces compared to random arrangements. For both open and closed structures, 128 number of obstacles is the highest obstacle density in which at least one path planner finds a feasible path. In this case, RRT fails, and other path planners determine a path successfully. Similar to the random obstacle arrangements, RRT finds a feasible path in lower obstacle densities.

For an open structure (Fig. 6(a)) with periodic obstacles, Dijkstra and A* perform similarly, though outperform PRM, with ~0.06s shorter run time, ~0.39mm lower RMSE, ~2.81mm lower offset, ~11 fewer number of turns and ~11.65° less roughness.

Fig. 5. Performance assessment of path planners in an environment with random point obstacles in a) open and b) closed structures

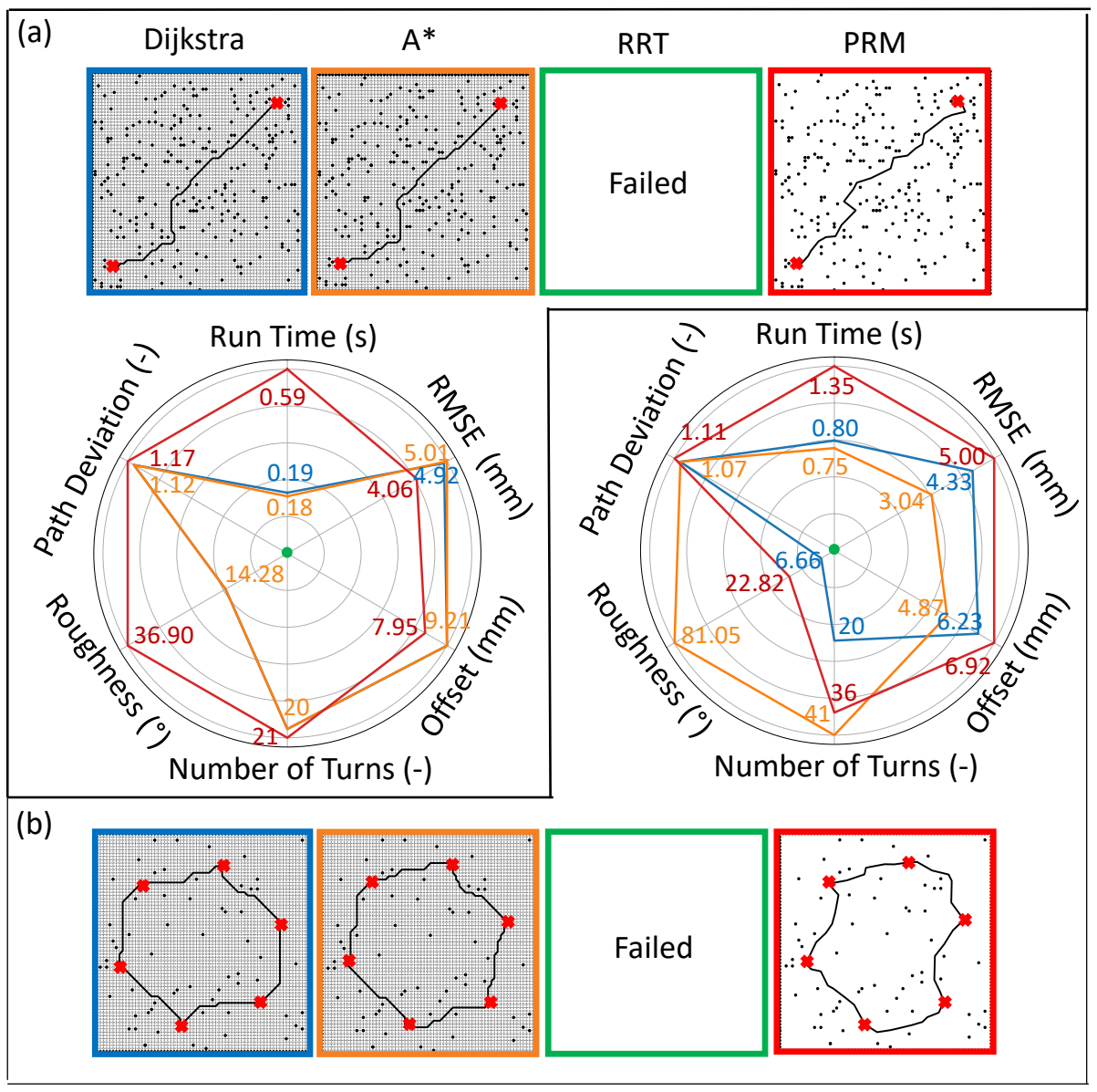

For a closed structure (Fig. 6(b)) with periodic obstacles, A* exhibits the shortest run time (0.82s), lowest RMSE (2.54mm) and offset (3.00mm), although Dijkstra has fewer turns (17) and the least roughness (5.31°).

Based on the resulting generated paths, roughness and number of turns represent the smoothness most accurately.

Fig. 6. Performance assessment of path planners in an environment with periodic point obstacles in a) open and b) closed structures

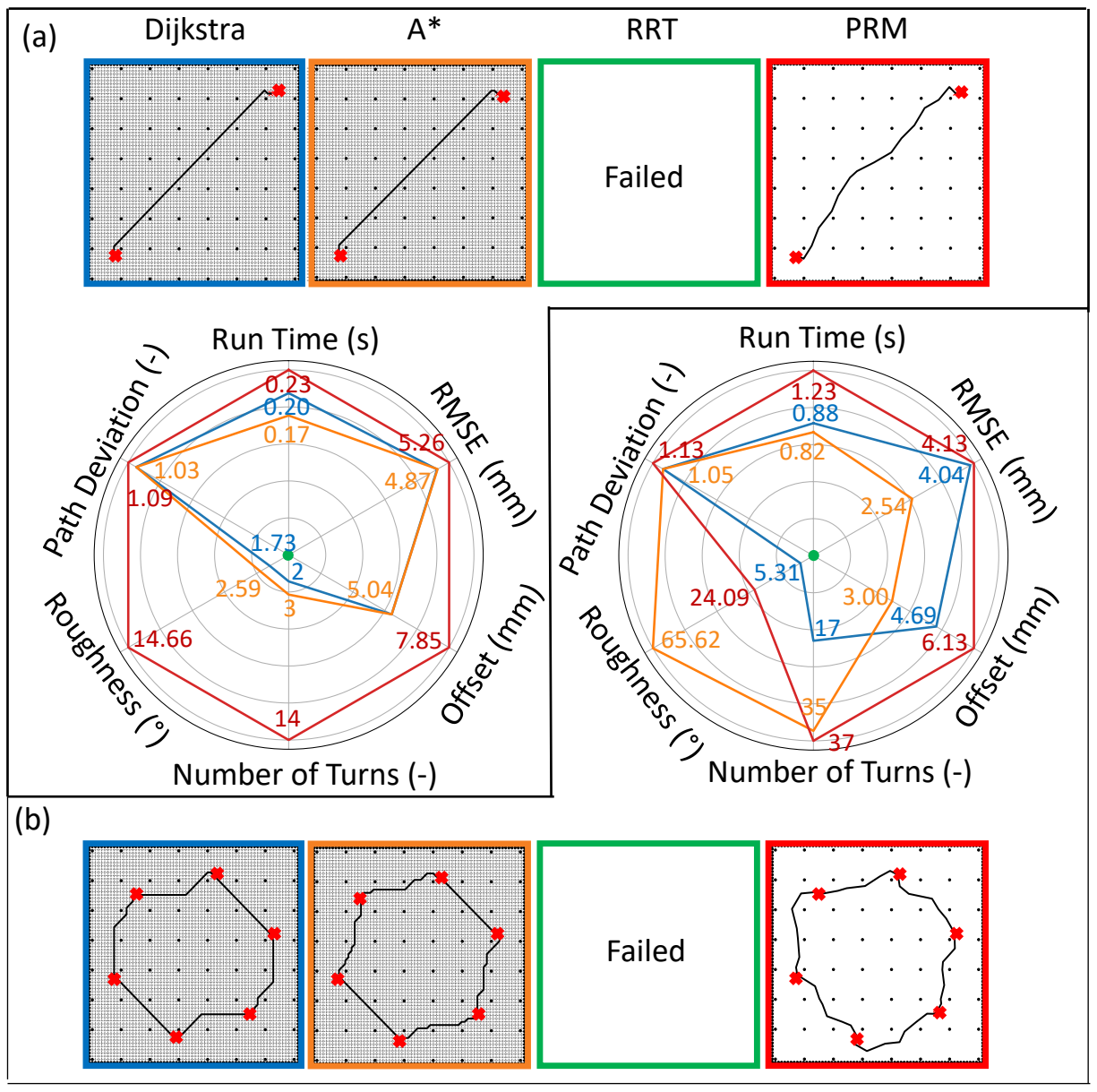

To compare the path planners in performance metrics, each path planner with their corresponding performance assessment metric is demonstrated in Fig. 7. The lower the metrics, the higher the corresponding performance. In contrast to Fig. 5 and 6, the maximum value of each performance metric is fixed in Fig. 7 among all path planners to simplify the comparison for each type of structure.

In open structures, overall, Dijkstra and A* have the shortest run time and lowest offset, as well as the least roughness and number of turns, although PRM has the lowest RMSE.

In closed structures, on average, all three path planners perform similarly in run time. Additionally, Dijkstra and A* have the lowest RMSE, offset, and fewer number of turns. Also, Dijkstra exhibits the best performance in the roughness.

Based on the performance study of path planners, Dijkstra performed the best, followed by A*, PRM, and RRT.

Fig. 7. Summary of performance assessment of (a) Dijkstra, (b) A*, and (c) PRM PPs in all obstacle arrangements and structures

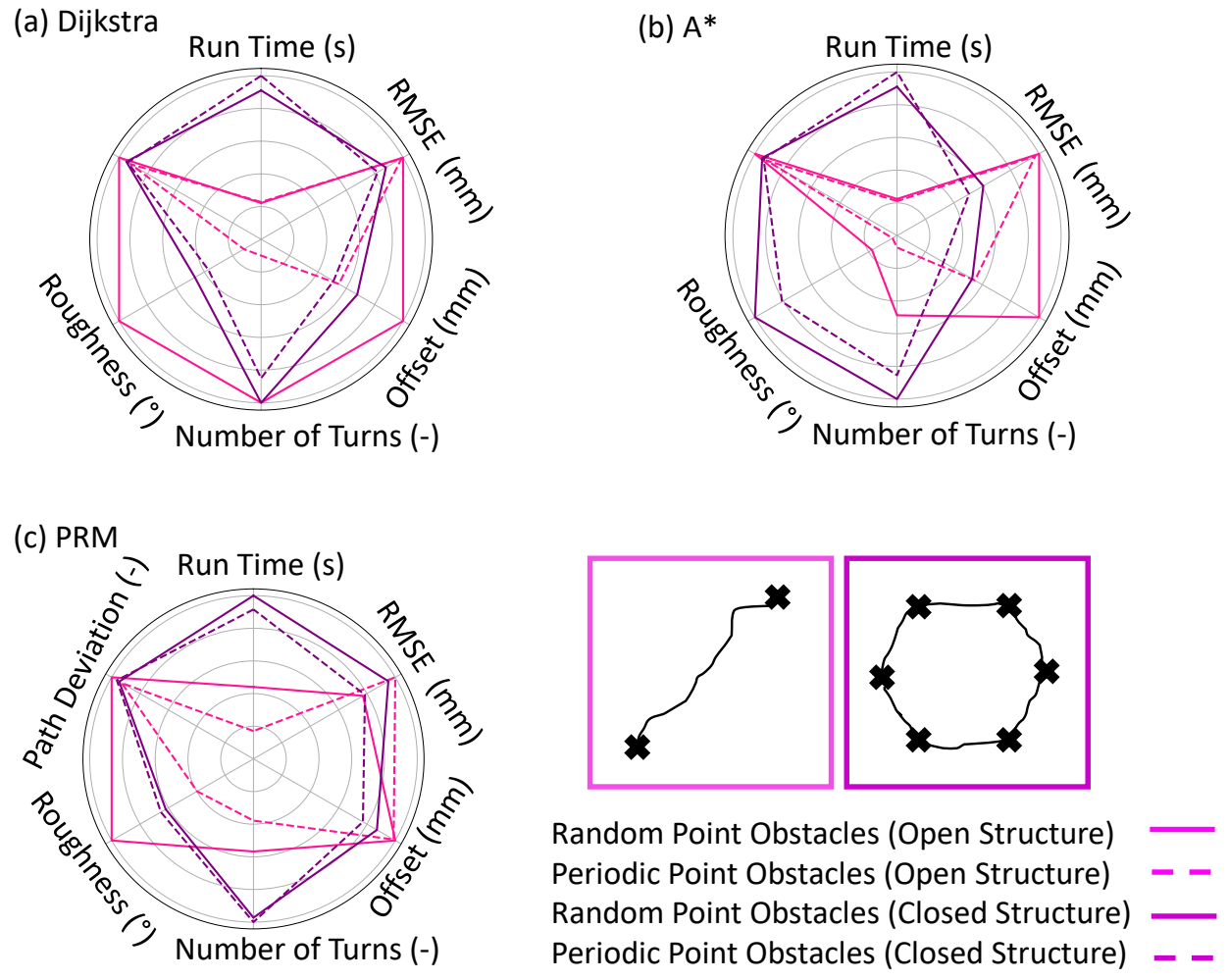

Among the performance metrics, path deviation has the least importance in making structural distinctions as values are closely similar. Moreover, based on the resulting generated path, roughness and number of turns represent the smoothness of the path in terms of angle change most accurately. Additionally, offset and RMSE both represent the length deviation of the generated path. Therefore, three performance metrics including roughness, number of turns, and RMSE provide the best information of the generated path with structural consideration and can be sufficient for assessment.

## V. Conclusion and Future Work

In this paper, an online path generation framework (PGF) for robotic additive manufacturing (AM) of structures is proposed for the first time. The PGF generates the toolpath of the structure, while it remains aware of the surrounding environment (i.e., obstacles) by utilizing four path planning (PP) algorithms.

Obstacle arrangements including periodic and random are considered for two types of structures (closed and open). Performance metrics including structural (path roughness, turns, offset, RMSE, deviation) and computational (run time) categories are developed. The feasibility of finding a path is assessed for the case of most challenging environments (i.e., most dense by suturing the number of obstacles) in order to probe the upper bounds of the path planners (until at least one succeeds). The effectiveness of the proposed structural performance metrics is evaluated individually and comparatively, and most essential metrics necessary for evaluation of toolpath of the resulting structures are prescribed.

The results demonstrate that in open structures, Dijkstra and A* outperform other path planners in run time. Similarly, A* outperforms other path planners in roughness, number of turns, offset and RMSE metrics. In closed structures, Dijsktra, A*, and PRM perform similar to one another in run time. Additionally, Dijkstra demonstrates the best performance (i.e., least values) in roughness, offset and RMSE, similar to, Dijkstra and A* in terms of number of turns. Although based on observations of the generated path, PRM demonstrates a favorably better performance in avoiding sharp corners and turns in closed structures, it undesirably has the highest offset and RMSE. Therefore, it can be concluded that Dijkstra and in the next steps, A*, can be revised to enhance the sharp corners using smoothing internal techniques such as internal smoothing functions of robotic manipulators. Based on the overall performance metrics, Dijkstra demonstrates the most promising outcomes and utility for implementation in both open and closed structures, followed by A* and PRM, where RRT rapidly fails in dense obstacle environment.

Additionally, this work highlights that roughness, number of turns, and RMSE are the most essential structural performance metrics to assess the PP algorithms in robotic AM. However, the performance metrics provide the tool for performance assessment. The objectives with regard to structural and geometric requirements will depend on the exact application.

The proposed approach for path generation in robotic AM (i.e., 3D-printing) eliminates the need for the *A priori* design of a structure (i.e., offline path generation) in an unknown environment beforehand. The PGF only requires a set number of target points, that can be prescribed by the user such as fixed start and end vertices for an open straight wall structure or the main vertices of a closed structure.

## Acknowledgment

## References

[1] R. Moini, "Perspectives in architected infrastructure materials," *RILEM Technical Letters*, vol. 8, pp. 125–140, Jul. 2023, doi: 10.21809/rilemtechlett.2023.183.

[2] D. Daneshvar, M. Rabiei, S. Gupta, A. Najmeddine, A. Prihar, and R. Moini, “Geometric Fidelity of Interlocking Bodies in Two-Component Robotic Additive Manufacturing,” in *Fourth RILEM International Conference on Concrete and Digital Fabrication*, D. Lowke, N. Freund, D. Böhler, and F. Herding, Eds., Cham: Springer Nature Switzerland, 2024, pp. 134–141.

[3] L. Tomholt, F. Meggers, and R. Moini, “3D-Printing Channel Networks with Cement Paste,” in *Fourth RILEM International Conference on Concrete and Digital Fabrication*, D. Lowke, N. Freund, D. Böhler, and F. Herding, Eds., Cham: Springer Nature Switzerland, 2024, pp. 74–82.

[4] N. Ralston, S. Gupta, and R. Moini, “3D-printing of architected calcium silicate binders with enhanced and in-situ carbonation,” *Virtual Phys. Prototyp.*, vol. 19, no. 1, 2024, doi: 10.1080/17452759.2024.2350768.

[5] S. Gupta and R. Moini, “Tough Cortical Bone-Inspired Tubular Architected Cement-Based Material with Disorder,” *Advanced Materials*, Dec. 2024, doi: 10.1002/adma.202313904.

[6] R. Moini, F. Rodriguez, J. Olek, J. P. Youngblood, and P. D. Zavattieri, “Mechanical properties and fracture phenomena in 3D-printed helical cementitious architected materials under compression,” *Mater. Struct.*, vol. 57, no. 7, p. 170, 2024, doi: 10.1617/s11527-024-02437-4.

[7] H. Shen, L. Pan, and J. Qian, “Research on large-scale additive manufacturing based on multi-robot collaboration technology,” *Addit. Manuf.*, vol. 30, pp. 1–10, Dec. 2019, doi: 10.1016/j.addma.2019.100906.

[8] M. E. Tiryaki, X. Zhang, and Pham Quang-Cuong, “Printing-while-moving: a new paradigm for large-scale robotic 3D Printing,” in *2019 IEEE/RSJ International Conference on Intelligent Robots and Systems (IROS)*, Macau, China, Nov. 2019, pp. 2286–2291. doi: 10.1109/IROS40897.2019.8967524.

[9] K. Zhang *et al.*, “Aerial additive manufacturing with multiple autonomous robots,” *Nature*, vol. 609, no. 7928, pp. 709–717, Sep. 2022, doi: 10.1038/s41586-022-04988-4.

[10] K. Dörfler *et al.*, “Additive Manufacturing using mobile robots: Opportunities and challenges for building construction,” *Cem. Concr. Res.*, vol. 158, Aug. 2022, doi: 10.1016/j.cemconres.2022.106772.

[11] S. Rescsanski, R. Hebert, A. Haghighi, J. Tang, and F. Imani, “Towards intelligent cooperative robotics in additive manufacturing: Past, present, and future,” *Robot. Comput. Integr. Manuf.*, vol. 93, pp. 1–28, 2025, doi: https://doi.org/10.1016/j.rcim.2024.102925.

[12] G. K. Al-Chaar *et al.*, “ERDC TR-21-3 ‘Automated Construction of Expeditionary Structures (ACES) : Materials and Testing,’” 2021. [Online]. Available: https://erdclibrary.on.worldcat.org/discovery.

[13] National Aeronautics and Space Administration (NASA), “3D-Printed Habitat Challenge,” 2019. [Online]. Available: www.nasa.gov

[14] M. Moini, J. Olek, J. P. Youngblood, B. Magee, and P. D. Zavattieri, “Additive Manufacturing and Performance of Architectured Cement-Based Materials,” *Advanced Materials*, vol. 30, no. 43, pp. 1–11, Oct. 2018, doi: 10.1002/adma.201802123.

[15] A. Paolini, S. Kollmannsberger, and E. Rank, “Additive manufacturing in construction: A review on processes, applications, and digital planning methods,” *Addit. Manuf.*, vol. 30, pp. 1–13, Dec. 2019, doi: 10.1016/j.addma.2019.100894.

[16] A. Prihar, S. Gupta, H. S. Esmaeeli, and R. Moini, “Tough double-bouligand architected concrete enabled by robotic additive manufacturing,” *Nature Communications* , vol. 15, no. 1, pp. 1–11, Dec. 2024, doi: 10.1038/s41467-024-51640-y.

[17] M. Rabiei and R. Moini, “Extrusion under material uncertainty with pressure-based closed-loop feedback control in robotic concrete additive manufacturing,” *Autom. Constr.*, vol. 180, Dec. 2025, doi: 10.1016/j.autcon.2025.106494.

[18] A. Kazemian, X. Yuan, O. Davtalab, and B. Khoshnevis, “Computer vision for real-time extrusion quality monitoring and control in robotic construction,” *Autom. Constr.*, vol. 101, pp. 92–98, May 2019, doi: 10.1016/j.autcon.2019.01.022.

[19] E. Shojaei Barjuei, E. Courteille, D. Rangeard, F. Marie, and A. Perrot, “Real-time vision-based control of industrial manipulators for layer-width setting in concrete 3D printing applications,” *Advances in Industrial and Manufacturing Engineering*, vol. 5, pp. 1–14, 2022, doi: https://doi.org/10.1016/j.aime.2022.100094.

[20] M. Rabiei, B. Gorse, and R. Moini, “Vision Alignment Mechanism for Feedback Control in Robotic Additive Manufacturing with Limited Degrees of Freedom,” *Under Review*, 2025.

[21] H. Fan *et al.*, “New era towards autonomous additive manufacturing: a review of recent trends and future perspectives,” *International Journal of Extreme Manufacturing*, vol. 7, no. 3, p. 032006, Jun. 2025, doi: 10.1088/2631-7990/ada8e4.

[22] Y. Jin, Y. He, G. Fu, A. Zhang, and J. Du, “A non-retraction path planning approach for extrusion-based additive manufacturing,” *Robot. Comput. Integr. Manuf.*, vol. 48, pp. 132–144, Dec. 2017, doi: 10.1016/j.rcim.2017.03.008.

[23] M. Ozcan, Y. Hergul, and U. Yaman, “Layer path updating algorithm based on Voronoi diagrams for material extrusion additive manufacturing,” *Int. J. Comput. Integr. Manuf.*, pp. 1–14, Jan. 2025, doi: 10.1080/0951192X.2025.2457110.

[24] M. Bi *et al.*, “Continuous contour-zigzag hybrid toolpath for large format additive manufacturing,” *Addit. Manuf.*, vol. 55, Jul. 2022, doi: 10.1016/j.addma.2022.102822.

[25] M. Geuy, J. Martin, T. Simpson, and N. Meisel, "Path Planning for Non-Planar Robotic Additive Manufacturing," in *Solid Freeform Fabrication 2023: Proceedings of the 34th Annual International Solid Freeform Fabrication Symposium – An Additive Manufacturing Conference*, 2023.

[26] H. Giberti, L. Sbaglia, and M. Urgo, "A path planning algorithm for industrial processes under velocity constraints with an application to additive manufacturing," *J. Manuf. Syst.*, vol. 43, pp. 160–167, Apr. 2017, doi: 10.1016/j.jmsy.2017.03.003.

[27] L. Han, W. Du, Z. Xia, B. Gao, and M. Yang, "Generative Design and Integrated 3D Printing Manufacture of Cross Joints," *Materials*, vol. 15, no. 14, Jul. 2022, doi: 10.3390/ma15144753.

[28] P. Nicholas, G. Rossi, E. Williams, M. Bennett, and T. Schork, "Integrating real-time multi-resolution scanning and machine learning for Conformal Robotic 3D Printing in Architecture," *International Journal of Architectural Computing*, vol. 18, no. 4, pp. 371–384, Dec. 2020, doi: 10.1177/1478077120948203.

[29] B. Felbrich, T. Schork, and A. Menges, "Autonomous robotic additive manufacturing through distributed model-free deep reinforcement learning in computational design environments," *Construction Robotics*, vol. 6, no. 1, pp. 15–37, Mar. 2022, doi: 10.1007/s41693-022-00069-0.

[30] K. E. Bekris, J. Doerr, P. Meng, and S. Tangirala, "The State of Robot Motion Generation," Oct. 2024, [Online]. Available: http://arxiv.org/abs/2410.12172

[31] E. W. Dijkstra, "A Note on Two Problems in Connexion with Graphs," *Numer. Math. (Heidelb).*, vol. 1, pp. 269–271, 1959.

[32] M. J. D Powell *et al.*, "The gradient projection method for nonlinear programming, pt. I, linear constraints," *Research Analysis Corp*, vol. 19, no. 2, pp. 874–890, 1968.

[33] S. M. LaValle, "Rapidly-exploring random trees : a new tool for path planning," *The annual research report*, 1998, [Online]. Available: https://api.semanticscholar.org/CorpusID:14744621

[34] L. E. Kavralu, P. Svestka, J.-C. Latombe, and M. H. Overmars, "Probabilistic Roadmaps for Path Planning in High-Dimensional Configuration Spaces," 1996.

[35] A. Sakai *et al.*, "PythonRobotics: a Python code collection of robotics algorithms," 2018. [Online]. Available: https://atsushisakai.github.io/

[36] Atsushi Sakai, "Python sample codes and textbook for robotics algorithms," GitHub. Accessed: Feb. 25, 2026. [Online]. Available: https://github.com/AtsushiSakai/PythonRobotics

[37] H. Y. Hsueh *et al.*, "Systematic comparison of path planning algorithms using PathBench," *Advanced Robotics*, vol. 36, no. 11, pp. 566–581, 2022, doi: 10.1080/01691864.2022.2062259.